\documentclass[letterpaper, 10 pt, conference]{ieeeconf}  

\IEEEoverridecommandlockouts                              

\usepackage{graphics} 
\usepackage{epsfig} 
\usepackage{mathptmx} 
\usepackage{times} 
\usepackage{amsmath} 
\usepackage{amssymb}  
\usepackage{stmaryrd}
\usepackage{svg}
\usepackage{float}
\usepackage{booktabs} 
\usepackage{adjustbox}
\usepackage{siunitx} 

\usepackage[linesnumbered,ruled,noend]{algorithm2e}
\usepackage{graphicx}
\usepackage{wrapfig}
\newcommand{\bs}[1]{\boldsymbol{#1}}

\usepackage{adjustbox}
\usepackage{booktabs} 
\usepackage{multirow}

\newcommand{\blue}[1]{{\color{blue} #1}}

\title{\LARGE \bf
Interactive Perception for Deformable Object Manipulation
}

\author{Zehang Weng$^{*,1}$, Peng Zhou$^{*,\dag,2}$, Hang Yin$^{1}$, Alexander Kravberg$^{1}$, \\
Anastasiia Varava$^{1}$,  David Navarro-Alarcon$^{2}$ and Danica Kragic$^{1}$
\thanks{$^*$Authors with equal contribution. $^\dag$ Corresponding author. 
}
\thanks{$^{1}$The authors are with CAS/RPL, KTH, Stocholm,  Sweden. {\tt\small \{zehang,hyin,okr,varava,dani\}@kth.se}}%
\thanks{$^{2}$The authors are with The Hong Kong Polytechnic University, KLN, Hong Kong. {\tt\small \{jeffzhou@hku.hk,  dnavar@polyu.edu.hk\}}}%
}

\begin{document}

\maketitle
\thispagestyle{empty}
\pagestyle{empty}

\begin{abstract}
Interactive perception enables robots to manipulate the environment and objects to bring them into states that benefit the perception process. Deformable objects pose challenges to this due to significant manipulation difficulty and occlusion in vision-based perception. In this work, we address such a problem with a setup involving both an active camera and an object manipulator. Our approach is based on a sequential decision-making framework and explicitly considers the motion regularity and structure in coupling the camera and manipulator. We contribute a method for constructing and computing a subspace, called Dynamic Active Vision Space (DAVS), for effectively utilizing the regularity in motion exploration. The effectiveness of the framework and approach are validated in both a simulation and a real dual-arm robot setup. Our results confirm the necessity of an active camera and coordinative motion in interactive perception for deformable objects. 
\end{abstract}

\section{Introduction}

Interactive Perception (IP) exploits various types of forceful interactions with the environment to facilitate perception ~\cite{bohg2017interactive}. Specifically, interaction allows for jointly considering acquired sensor information and actions taken over a time span. Despite the high dimensionality of the augmented space, the causal relation between sensation and action gives rise to structure for perceiving environment properties in a predictive and dynamical manner. Such structure is called Action Perception Regularity~\cite{bohg2017interactive}, and is believed to be the key for IP to reveal richer signals, which is impossible for passive and one-shot perception. IP has been shown effective in exploring and exploiting object and environment properties, with the main focus on rigid and articulated objects ~\cite{cheng2018reinforcement,novkovic2020object,martin2014online,katz2008manipulating,katz2014interactive}.

\begin{figure}[tbp]
  \begin{center}
    \includegraphics[width=0.9\linewidth]{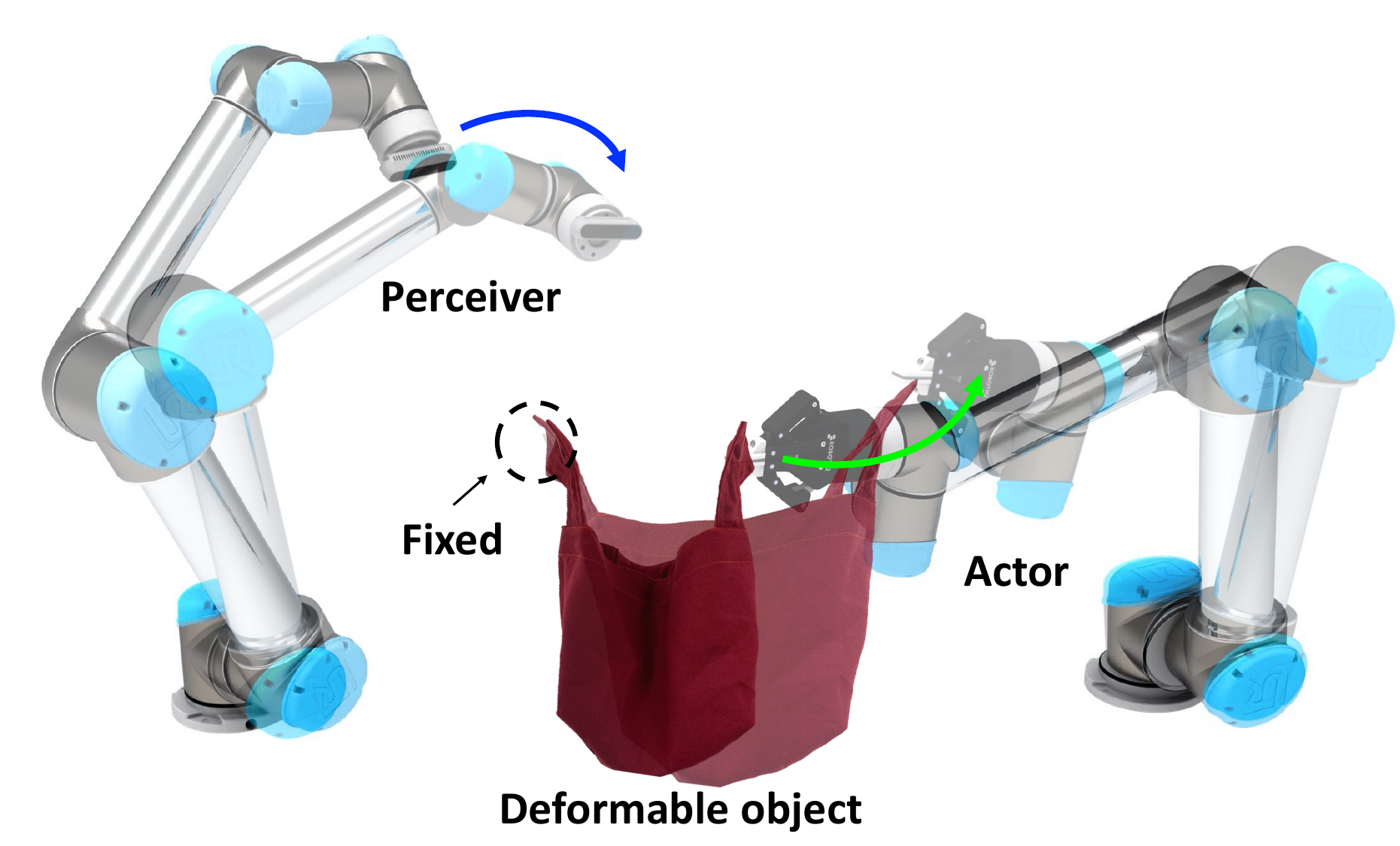}
  \end{center}
  \vspace{-0.2cm}
  \caption{An example of Interactive Perception. The perceiver (camera) is moved to a new viewpoint while the actor (end-effector) opens the bag for better perception of in-bag object.
  }
  \label{fig_ma_bo}
  \vspace{-0.5cm}
\end{figure}

Deformable object manipulation (DOM) adds to the complexity of this identified interactive perception problem (e.g., clothes or bags manipulation). The space where the perception process resides in is further enlarged with much more degrees-of-freedom (DOFs) of the deformable materials. Establishing the predicative relation between action and sensation is challenged by an underactuated system. Moreover, significant occlusions may present due to  material flexibility and fail an interaction strategy without an actively controlled camera. An example is shown in Fig.~\ref{fig_ma_bo}, where an active camera called \textit{perceiver} is necessitated for a better observation of an object in a bag while the bag is opened by another manipulator called \textit{actor}. This will again add more DOFs to the problem. Lastly, deformable environments might entail a perception process consisting of multiple steps: the robots need to take actions for better perception which in turn bases a better action decision to take. How to effectively coordinate the perceiver and the actor to shorten this interactive process can thus be a key to efficient perception.

To this end, we propose to address interactive perception for deformable objects in a partially observable Markov Decision Process (POMDP) framework. We focus on bag-like objects and a dual-arm agent setup with an active camera similar to Fig.~\ref{fig_ma_bo}. Our aim is to study the feasibility of solving DOM for IP under the POMDP formulation and whether active vision can facilitate the solution. 
To leverage the regularity in coupling the perceiver and the actor, our approach proposes to dynamically construct a motion subspace, Dynamic Active Vision Space (DAVS), based on object features for an efficient search. 
Specifically, we introduce a manifold-with-boundary formulation to characterize a compact camera action subspace to constrain the potential next view. We provide detailed algorithms to calculate the manifold and integrate it into a general action search framework such as reinforcement learning (RL) in solving the POMDP. 
The simulation results highlight the importance of active vision capability from the controlled camera and the superior performance due to DAVS compared to baseline methods. DAVS shows its strength in generalizing to unseen dynamics and shapes. Finally, we show the proposed method is robust for a successful transfer to the real hardware with a moderate fine-tuning on a dual-arm setup.

In summary, our main contributions are:
\begin{itemize}
\item We propose a formulation involving both an active camera and a highly deformable object, which to our knowledge is the first in the context of interactive perception.
\item We present a novel method that constructs the camera action space as a manifold with boundary (DAVS), based on Structure of Interest (SOI).
\item We conduct extensive studies in both simulation and real-world experiments to demonstrate the effectiveness of our methods in challenging DOM scenarios, We show that our method generalizes well to objects with different dynamical properties and unseen shapes.
\end{itemize}


\section{Related Work} 

\begin{figure*}[tb]
\centering
\includegraphics[width=0.95\linewidth]{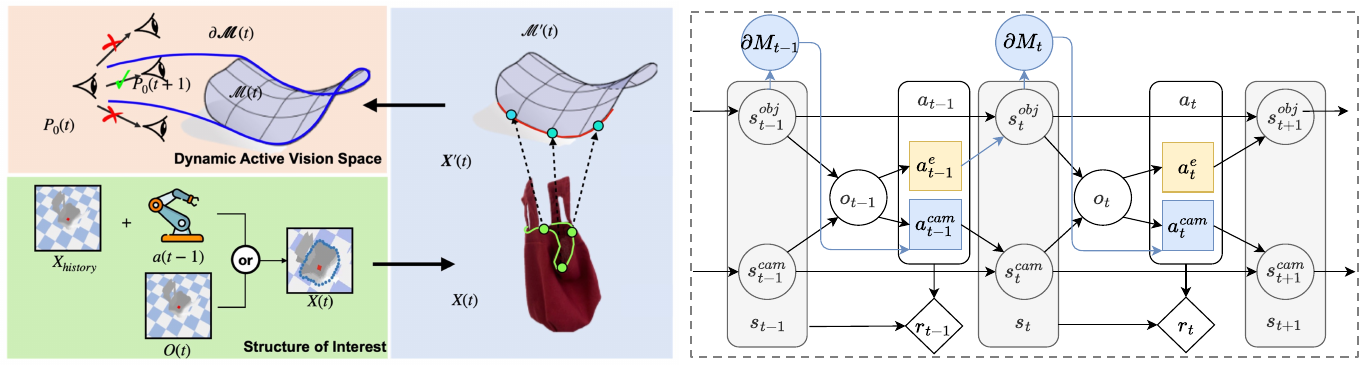}
\DeclareGraphicsExtensions.
\vspace{-0.3cm}
\caption{\small{[Left] Illustration of the proposed framework --- a subspace of camera action is constructed and represented with manifold with boundary, accounting for the coupling with end-effector motion via the structure of interest (SOI) on deformable objects. [Right] Illustration of state action transition of the proposed framework.}}
\vspace{-0.7cm}
\label{fig_framework}
\end{figure*}

Interactive Perception and manipulation of deformable objects ~\cite{yin2021modeling,arriola2020modeling,zhu2021challenges,herguedas2019survey} are commonly studied as two separate problems in the robotics community. In this work, we aim to develop a framework that allows for a more effective perception in a manipulation task performed on a deformable object.
Our work relates to concepts such as hand-eye coordination ~\cite{doi:10.1177/0278364917710318}, visual servoing ~\cite{Kragic02surveyon}, active perception ~\cite{aloimonos1988active,bajcsy2018revisiting,yuille1992active,ballard1991animate,rivlin2000control,zhang2023affordance}  and interactive perception ~\cite{bohg2017interactive}, but none of these fully addresses all the three aspects in a single framework: a deformable object, moving camera, and a robot interacting with the object. 

Recent work on learning policies in the context of IP focuses mostly on rigid objects and RL ~\cite{cheng2018reinforcement,novkovic2020object}. Works considering deformable objects and a moving camera reside commonly to active vision and do not address the interaction aspect ~\cite{sock2017multi,wenhardt2007active,vazquez2001viewpoint,van2012maximally}. RL is again employed in the recent work of ~\cite{gartner2020deep} but the focus is kept on resolving the observation for better human pose estimation with no interactive aspect. IP in the context of deformable objects has mainly been considered using a static camera ~\cite{yin2021modeling}. The most important aspect of these works is to address the dimensionality in terms of perception using a latent representation and then devising action planning in a lower-dimensional space ~\cite{lippi2020latent,zhou2021lasesom,manuelli2019kpam}. Notably, advancements in complex 3D deformable bag manipulation have been presented ~\cite{weng2021graph,bahety2022bag, xu2022dextairity,chen2023autobag, zhou2024bimanual}. These works have shown significant progress in the domain but do not incorporate an active movable camera for task achievement. The recent benchmarks that build upon advanced simulation engines again focus primarily on a static camera setup ~\cite{lin2020softgym,antonova2021dynamic}. 

The work in ~\cite{doi:10.1177/0278364917710318} learns hand-eye coordination for grasping, using deep learning to build a grasp success prediction network, and a continuous servoing mechanism to use this network to continuously control a robotic manipulator. Similarly to our approach, it uses visual servoing to move into a pose that maximizes the probability of success on a given task. However, the application is bin-picking, using a static camera, and no deformable objects are considered. As per technical methods, we also follow a decision-making formulation and reinforcement learning as ~\cite{cheng2018reinforcement,novkovic2020object}. Our work differs by contributing a factorization of the policy action space and algorithms that leverage the dependency between the camera and object for efficient search. The proposed structure is shown to be critical to addressing challenges faced by ~\cite{cheng2018reinforcement} in deformable object scenarios. 

In summary, we contribute a formulation of an interactive perception approach that relies on a manifold-with-boundary representation to address the high-dimensionality of the perception-action problem and efficiently encode the coupling between the perceiver, deformable object, and the actor.

\section{Methodology}

We formulate interactive perception involving deformable objects as a POMDP $({S}, {O}, {A}, {T}, {R}, \gamma)$, with state $s \in {S}$ denoting the configuration of robotic end-effector, camera, and object, and corresponding observation as $o \in {O}$. State transition model \({T}\left(s^{\prime} \mid s, a\right)\) characterizes the probability of transitioning to state \(s^{\prime}\) from taking action \(a \in {A}\) under state \(s\). The action space is comprised of desired camera and end-effector poses, represented as a Cartesian product ${A} = {A}_{cam} \times {A}_{e}$.  \({R}(s) \in \mathbb{R}\) is a state reward function, and \(\gamma \in[0,1)\) is the discount factor.

Our core idea to tackle the problem is constructing a subspace of ${A}$ for efficient action search and maximizing the accumulated reward. An overview of the framework components is depicted in Fig.~\ref{fig_framework}. Specifically, the subspace is built upon certain object features, which are called the structure of interest (SOI). The geometry of this structure is coupled with the end-effector action. The structure is used to generate the temporally varying boundary of a manifold, which is called dynamic active vision space. In the end, the camera actions are projected to the constructed space in a reinforcement learning process. We detail the mathematical representation of the manifold and the development of each component in the following sections.

\subsection{Action Space Factorization and Manifold with Boundary}




Given the IP task in Fig.~\ref{fig_ma_bo}, we need to propose the camera action $a^{cam}_{t}$ and end-effector action $a^{e}_{t}$ given the observation $o_{t-1}$. A naive solution is to factorize the action space as $\pi(a^{cam}_{t}, a^{e}_{t}|o_{t-1}) = p(a^{cam}_{t}|o_{t-1}) p(a^{e}_{t}|o_{t-1})$ with the assumption of conditional independence, which however results in high dimensionality for action exploration. In this work, we consider that the camera action space should also depend on the previous end-effector action, resulting in $\pi(a^{cam}_{t}, a^{e}_{t}|o_{t-1}) = p(a^{cam}_{t}|a^{e}_{t-1},o_{t-1}) p(a^{e}_{t}|o_{t-1})$. This implies the possibility of sampling $a^{cam}_{t}$ from a subspace depending on $a^{e}_{t-1}$, thereby enabling efficient exploration of the original action space. We use the \textit{manifold with boundary} $M$ \cite{lee2010introduction} as a tool to describe this subspace and its construction.

The introduction of ${M}$ essentially imposes constraints, and thus structure/regularity, on the original camera action space.  To this end, we augment the POMDP $({S}, {O}, {Z}, {T}, {R}, \gamma)$, such that $z \in {Z} = {M} \times {A}^{e}$ is the constrained action space subjecting to ${M}$. The new constrained IP problem can be written as

\begin{equation}
\begin{split}
	\max _{\bs{\theta}} \mathbb{E}_{\bs{s}_{t}, \bs{z}_{t}^{cam}, \bs{z}_{t}^{e} }\left[\sum_{t=0}^{T} \gamma^{t} {R}\left(\boldsymbol{s}_{t}, \bs{z}_{t}^{cam}, \bs{z}_{t}^{e} \right)\right], \quad\\
	s.t. \quad 
	 \bs{z}_{t}^{cam} \in \bs{M}(\bs{z}_{t-1}^{e}): \{\text{Int}{M}, \partial{{M}}\}
\end{split}
\end{equation}
with $\bs{\theta}$ as the learning parameters of model $\bs{M}$ and $\pi(\bs{z}^{e}|\bs{o})$.

\subsection{Manifold with Boundary-based DAVS}
The manifold with boundary specifies a compact subspace of the original camera action space. We adopt a half spherical space as the original space, which is often assumed in active perception research ~\cite{calli2018active,gibbs2019active,gartner2020deep}. Specifically, the spherical space is defined by viewpoint centroid, $V_0$ and a viewpoint radius $r$, 
with both as known parameters: $S^{2}(r)=\{\bs{p} \in \mathbb{R}^{3} \mid \|\bs{p}-V_0\|=r, z \geq 0\}$.

We seek to identify a subspace of $S^{2}$ to exploit the regularity of camera action space. This subspace is called Dynamic Active Vision Space (DAVS) for its dependency on the observation and end-effector action at each time step. To model such a dependency, we propose to use object features as an intermediary which are in turn influenced by end-effector actions. Specifically, we consider a set of key-points \(\mathbf{X}(t)=\left[X_{1}(t), X_{2}(t), \cdots, X_{N}(t)\right]^{T}\) with \(X_{i}(t) \in \mathbb{R}^{1 \times 3}\) as a representation of task-dependent features, such as garment opening and handle loop in Fig. \ref{fig_framework}, which are named as structure of interest (SOI). SOI can be acquired by using key-point extraction techniques ~\cite{manuelli2019kpam,manuelli2020keypoints} or predicative models ~\cite{sanchez2020learning,weng2021graph}. 

To focus on the contributed method, we assume annotated key-points that can be readily retrieved from a model $f_{SOI}(\bs{o}_t)$.
We expect SOI to resemble a single loop structure for the computation in subsequent implementation. This can be satisfied by design in the cases of, for instance, vertices on bag openings. To ensure such a structure, we process raw SOI key-points $\mathbf{X}(t)$ by first projecting them on a middle sphere with a radius $r^{\ast} = max(\| X_i(t)-V_0\|)$, with the projected points defined as:
\begin{equation}
\begin{split}
    \bar{\mathbf{X}}(t)  &=   V_0 +
    \frac{ (V_0 - X_i(t))r^{\ast}}{ \| X_i(t)-V_0\|} \\
    &\mid X_i(t) \in \{X_1(t), X_2(t), \ldots, X_N(t)\} = \mathbf{X}(t)
\end{split}
\end{equation}
Selecting the projected points $\mathbf{X}^{\ast}(t)$ on the convex hull of $\bar{\mathbf{X}}(t)$, we obtain a SOI polygon with desired loop structure.

\begin{figure*}[htbp]
\centering
\includegraphics[width=0.9\linewidth]{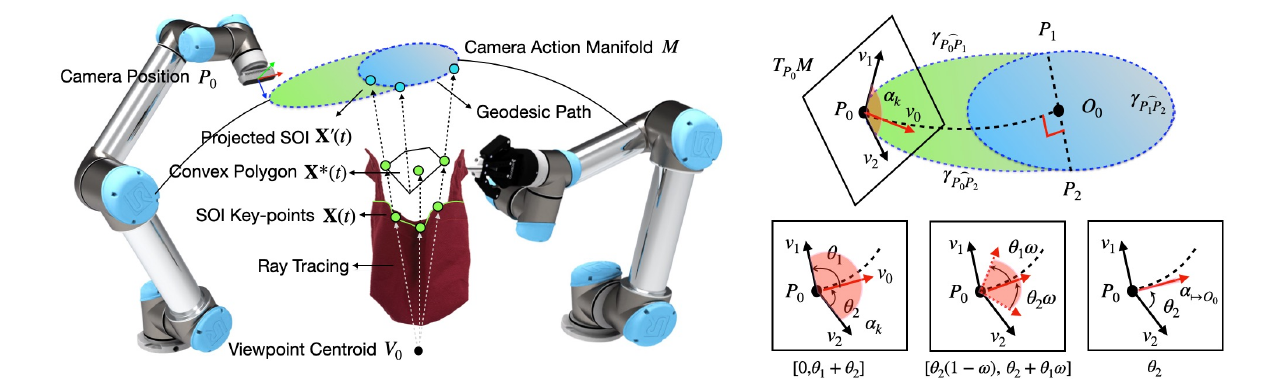}
\DeclareGraphicsExtensions.
\vspace{-0.3cm}
\caption{Illustration of the process of dynamic active vision space (DAVS) generation. [Left] We extract the SOI points at each time step $t$ and convert them to projected 3D SOI (blue) points through ray tracing, on the original camera action manifold. 
[Right top] The manifold with boundary based on projected SOI and current camera position by Algorithm \ref{alg_davs} [Right bottom] Parameterized exploration space.}
\label{fig_davs}
\vspace{-0.6cm}
\end{figure*}

We construct DAVS based on this SOI polygon to compute a subset of the original camera action space $S^2$.
A subset of $S^2$ at each time step is built by projecting key-points $\mathbf{X}^{\ast}(t)$ to $\mathbf{X}'(t)$ with a ray-tracing model, as shown in Fig. \ref{fig_davs}:

\begin{equation}
\begin{split}
    \mathbf{X}'_i(t)  &= V_0 + \frac{ (V_0 - X^{\ast}_i(t))r}{ \| X^{\ast}_i(t)-V_0\|} \\
    &\mid X^{\ast}_i(t) \in \{X^{\ast}_1(t), X^{\ast}_2(t), \ldots, X^{\ast}_n(t)\} = \mathbf{X}^{\ast}(t)
    \label{equ_SOI}
    \end{split}
\end{equation}

By connecting the $n$ projected SOI points in their respective order with $N$ geodesic paths, we generate a manifold ${M}'$ with boundary $\partial {M}'$ (which is part of $S^2$) consisting of the piecewise geodesic paths. In our implementation, we represent the geodesic path as a discrete set of path points within a certain density range to accelerate the computation process ~\cite{zhou2021lasesom}.

\begin{algorithm}[!ht]
\label{alg_rl_all}
\caption{Action Exploration with DAVS} 
\LinesNumbered
\KwIn{
Observation, $\bs{o}$;
Camera position, $P_0$;
Viewpoint centroid, $V_0$;
Viewpoint radius, $r$.

}
\KwOut{
Camera action policy, $\bs{\alpha}_k^{cam}$;
End-effector action policy, $\bs{\alpha}_k^{e}$ \;
}
\For {\textit{each episode}}
{
	Initial feasible state $\bs{s}_0$ \;
	Initial parameter setting for active vision \;
	\For {\textit{each time step $k$}}
	{
		SOI Modeling $\mathbf{X}(t) \leftarrow f_{SOI}(\bs{o}_{k-1})$ \;
		Sample policy end-effector action \( \bs{\alpha}_{k}^{e} \sim \pi(\cdot \mid \bs{o}_{k-1})\) \;
		Based on $\mathbf{X}(t)$, $P_0$, $V_0$, and $r$, compute DAVS using Alg. (\ref{alg_davs}) to get ${M}$ with its boundary $\partial {M}$ \;
		Compute direction vectors $\bs{v}_1$ and $\bs{v}_2$ based on geodesic paths on  $\gamma_{P_0P_1}$ and $\gamma_{P_0P_2}$ \;
		Sample policy camera action \( \bs{\alpha}_{k}^{cam} \sim \pi(\cdot \mid \bs{o}_{k-1}, \bs{v}_1, \bs{v}_2, \bs{v}_0, \omega)\) \;
		Apply the control camera action $\bs{\alpha}_{k}^{cam}$ to the environment \;
		Construct action tuple $\bs{\alpha}_k=(\bs{\alpha}_{k}^{e},\bs{\alpha}_{k}^{cam})$ \;
		Get next observation $\bs{o}_{k+1}$ and reward $r_k$ from the environment \;
		Provide the transition tuple $(\bs{o}_{k-1}, \bs{\alpha}_k, \bs{o}_{k}, r_k)$ to the RL algorithm \;
	}
}
\end{algorithm}

\begin{algorithm}[!ht]
\label{alg_davs}
\caption{Dynamic Active Vision Space (DAVS) Generation} 
\LinesNumbered
\KwIn{
SOI set (ring vertices), $\mathbf{X}(t)$;  
Camera position, $P_0$;
Viewpoint centroid, $V_0$;
Viewpoint radius, $r$.
}
\KwOut{Manifold ${M}$ with boundary $\partial {M}$ representing the reshaped camera action space}
Compute $\mathbf{X}'_i(t)$ with ray tracing model using Eq. (\ref{equ_SOI}) \;
Generate refined active vision space, ${M}'(t)$ and its boundary $\partial {M}'(t)$ \;
Compute centroid $O_0$ using the trust-region method with Eq. (\ref{equ_trust_manifold}) \;
Compute $P_1$ and $P_2$ s.t. $ Geodesic_{P_1P_2} \perp Geodesic_{P_0O_0}$ \;
Form manifold with boundary ${M}$ s.t. boundary $\partial M = \{ p \mid p \in \gamma_{\overset{\frown} {P_0P_1}}, \gamma_{\overset{\frown}{P_1P_2}}, \gamma_{\overset{\frown}{P_2P_0}} \} $
(each $p$ has a neighborhood homeomorphic to an open subset of \textit{closed n-dimensional upper half-space} \(\mathbb{H}^{n}\).
\end{algorithm}
\vspace{-0.5cm}


A DAVS for the current camera pose is constructed by finding a centroid on ${M}'(t)$, which is calculated as the Karcher mean of the boundary points $\mathbf{X}'(t)$, see Fig. \ref{fig_davs}. 
Formally, \(({M}', d)\) defines a complete metric space and $d$ denotes the geodesic distance in this metric space. 
For any position \(p\) in \({M}'\), the Fréchet variance \(\Psi\) is defined to be the sum of squared distances from \(p\) to the \(X'_{i}\) 
and is minimized as:
\begin{equation}
	f_{O_0}(p)= \underset{p \in M}{\arg \min } \Psi(p) = \underset{p \in M}{\arg \min } \sum_{i=1}^{N} d^{2}\left(p, X_{i}\right)
\end{equation}
To efficiently solve this problem, we follow ~\cite{centroid_python_solver} to transform it into a trust-region subproblem on \(T_{p_{k}} {M}'\). 
Given the cost function \(f_{O_0}: {M}' \rightarrow \mathbb{R}\) and a current iterate \(p_{k} \in {M}'\), we use retraction \(R_{p_{k}}\) to locally map the minimization problem for \(f\) on \({M}\) into a minimization problem for its pullback $\widehat{f}_{O_0}(p_{k}): T_{p_{k}{M}'}\rightarrow \mathbb{R}$, with which the trust-region problem reads:

\begin{equation}
\begin{split}
	\min _{\eta \in T_{p_{k}} {M}} \widehat{f}_{p_{k}}(\eta)&=f\left(p_{k}\right)+\left\langle\operatorname{grad} f\left(p_{k}\right), \eta\right\rangle \\
	&+\frac{1}{2}\left\langle H_{k}[\eta], \eta\right\rangle
	\text{s.t.} \enspace \langle\eta, \eta\rangle_{p_{k}} \leq \Delta_{k}^{2}
\end{split}
\label{equ_trust_manifold}
\end{equation}

For the symmetric operator \(H_{k}\) we use Hessian $\operatorname{Hess} f(x)[\xi]=\nabla_{\xi} \operatorname{grad} f(x)$. A trust region is specified by $\Delta_{k}$ for the validity of ambient space approximation. 

In addition to the centroid, we find two points $P_1$ and $P_2$ on $\partial {M}'(t)$, such that the geodesic path passing through them is locally perpendicular to the one passing through $P_0$ and $O_0$. 
As a result, the DAVS is obtained by enclosing the three paths, $\gamma_{\overset{\frown} {P_0P_1}}, \gamma_{\overset{\frown}{P_1P_2}}, \text{and} \gamma_{\overset{\frown}{P_0P_2}}$. 
The full process is summarized as Algorithm \ref{alg_davs}.
Additionally, we introduce a $\omega$ to parameterize the exploration space of the camera's action. As illustrated in Fig. \ref{fig_davs}, consider the tangent space $T_{P_0}M$ at the current camera position $P_0$ to a DAVS. Here, the local geodesic paths $\gamma_{\overset{\frown} {P_0P_1}}$, $\gamma_{\overset{\frown} {P_0P_2}}$, and $\gamma_{\overset{\frown} {P_0O_0}}$ are projected onto this tangent space, resulting in vectors $v_1$, $v_2$, and $v_0$.
Let us assume that angles $\theta_1$ and $\theta_2$ are formed between $v_1$ and $v_0$, and $v_2$ and $v_0$, respectively. By introducing a parameter $\omega \in [0, 1]$, we gain control over the camera action exploration within the range $[\theta_2 (1-\omega), \theta_2 + \theta_1 \omega]$ ($v_2$ as zero-degree axis). 
In our manuscript, we select the action space with $\omega = 1$ to provide a broader space for exploring effective policies in complex scenarios.
By reducing $\omega$, we can narrow the action sampling space and enforce a preference for alignment with $v_0$, which corresponds to the direction along the geodesic path $\gamma_{\overset{\frown} {P_0O_0}}$. Reducing $\omega$ to 0 causes the camera action policy to strictly follow the geodesic path $\gamma_{\overset{\frown} {P_0O_0}}$, leading directly to the centroid of DAVS. 

\subsection{Exploiting DAVS in Action Exploration}
The construction of DAVS provides a subspace that associates promising camera actions to task-relevant object features under the end-effector action. To this end, we propose to exploit this space to bias the search of camera action in IP, e.g., in reinforcement learning. As shown in Fig.~\ref{fig_framework}, we determine two tangent directions, $\bs{v}_1$ and $\bs{v}_2$, based on the boundary of DAVS at the current camera position. The camera action is sampled with a direction between $\bs{v}_1$ and $\bs{v}_2$. This biases the action exploration towards the DAVS and as such can boost search efficiency when object features indicate rewarding areas. We detail the entire process as Alg.~\ref{alg_rl_all} in the context of action sampling in reinforcement learning.




\begin{figure*}[htbp]
\centering
  \includegraphics[width=1\textwidth,height=5cm]{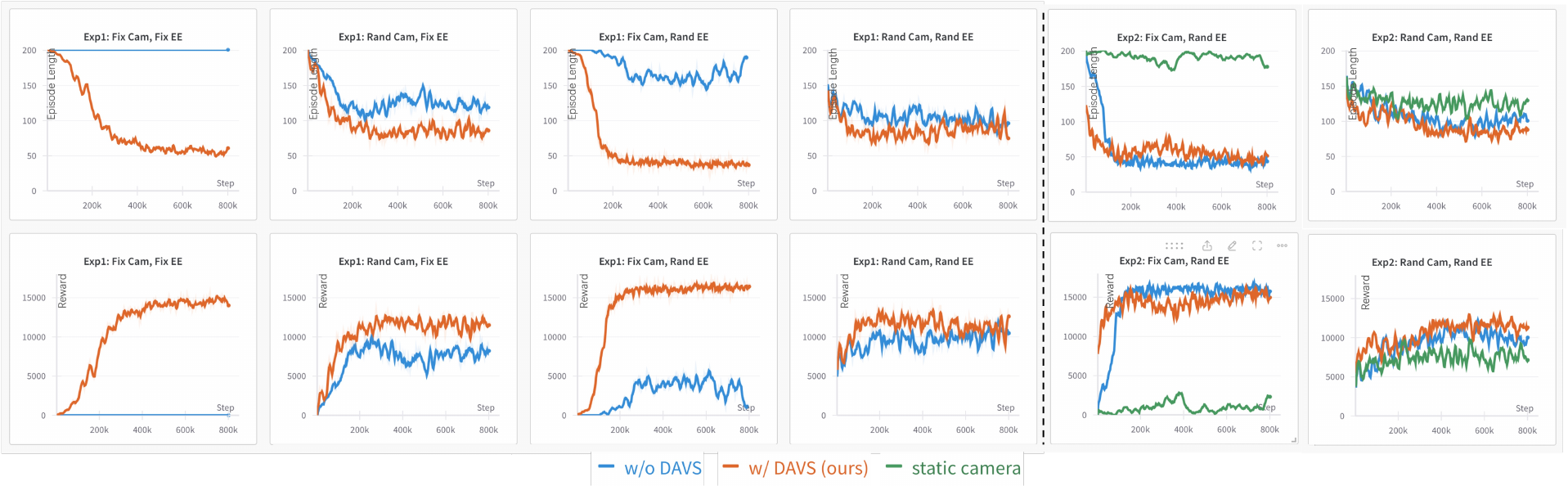}
   \vspace{-0.8cm}
\caption{
\small{Left: quantitative evaluation on CubeBagClean. The first row represents the \textbf{episode length} ($\downarrow$) for solving 4 tasks across different combinations of random (Rand) and fixed (Fix) camera (Cam) and end-effector (EE) birth modes. The second row shows the \textbf{total discounted rewards} ($\uparrow$).}
\small{Right: quantitative evaluation on CubeBagObst subtasks. The first two figures represent the \textbf{episode length} ($\downarrow$) and \textbf{reward} ($\uparrow$) for the scenario with fixed camera (Fix Cam) and random end-effector (Rand EE) birth modes. The third and fourth figures are \textbf{episode length} ($\downarrow$) and \textbf{reward} ($\uparrow$) with random camera (Rand Cam) and random end-effector (Rand EE) birth modes.}}
\label{exp2_clean_obst}
\vspace{-0.4cm}
\end{figure*}

\section{Simulation Experiments}
\label{sec:experiment}

\begin{figure}[tb]
\centering
    \includegraphics[width=0.48\textwidth]{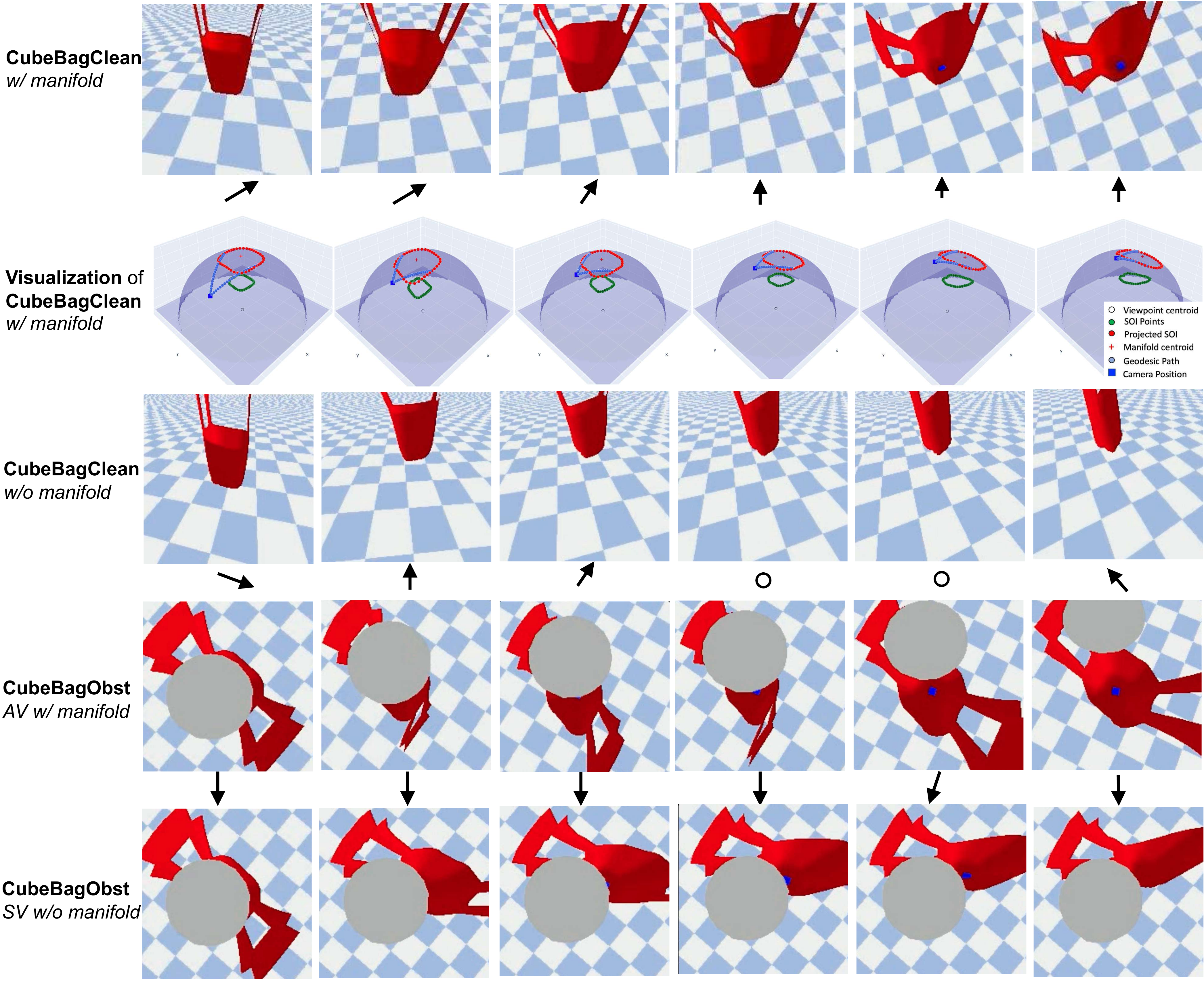}
\caption{\small{Visualization of the learned hand-eye policies. In each row except for the second row, we show the images of intermediate frames in an example episode. We also draw arrows $\uparrow$ beneath the images to illustrate how the camera moves. In the first and third rows sampled from the \textbf{CubeBagClean} case, we compare the performance between IP methods with and without DAVS. We can see that with DAVS, the camera is able to move right and upwards to find the cube in the bag, while the one without DAVS searches more randomly. In the bottom two rows from the \textbf{CubeBagObst} case, we compare the methods between IP methods with DAVS and a static vision method. In the IP setting, the camera bypasses the obstacle and finds a feasible solution, while the static vision method doesn't. This reveals the necessity of allowing active vision and active end-effector. The second row is the 3D manifold visualization corresponding to the first row (BlueDot: Camera; Green: 3D SOI; Red: Projected SOI; BlueCurve: Boundary of DAVS).}}
\label{case_qualitative}
\vspace{-0.4cm}
\end{figure}

\subsection{Environment Setup}
Our simulation environment is built on the basis of Pybullet ~\cite{coumans2016pybullet} which has been previously extended to handle various cloth-like deformable object manipulation tasks ~\cite{antonova2021dynamic,seita2021learning} under static viewpoint camera. We advance one step further and customize the environment to support the empirical evaluation of the proposed IP framework as shown in Fig. \ref{fig_davs}. 

We evaluate the performance of the proposed framework with DAVS in two environments: (1) CubeBagClean: environment contains a deformable bag with the left handle fixed in the air and a rigid cube placed inside; (2) CubeBagObst: CubeBagClean with an additional circular obstacle above the bag, providing an additional view occlusion. To ensure diversity of the initial position of the cube, it is spawned and released from a certain height above the bag, falling down into the bag freely.

In both simulation environments, we provide three types of observations: a 2D depth image, a 2D SOI heatmap, and the end-effector position. Based on the observations, the camera and end-effector manipulating the bag are allowed to act simultaneously, but restricted to a maximum of 200 action steps in one episode. Specifically, the end-effector can be arbitrarily moved while the active camera is traversing the half-sphere surface, similar to the CMU's panoptic massive camera grid ~\cite{joo2015panoptic} setting. The perceiver agent is controlled by pitch and yaw angles and stares at a constant focus point. To avoid collisions and ensure the simulation stability, we constrain the end-effector action with a bounding box and the camera yaw angle with a range.

\subsection{Evaluation Metrics}
We use CubeBagClean and CubeBagObst to assess the efficiency of using the proposed action space, in terms of succeeding in finding the cube. In CubeBagClean, we vary the birth modes of the camera and the end-effector to create 4 subtasks where either the camera or the end-effector can be spawned in a fixed or a random place. In CubeBagObst, we consider 2 subtasks with both fixed and random camera birth modes while the end-effector is always initialized in a random position. The fixed camera position is selected in a way that the bag interior is not immediately visible, so non-trivial camera trajectories are required to solve the task. In each task, our reward function encodes a goal of maximizing the visibility of the rigid cube and the opening ring. We use a potential-based reward formulation ~\cite{ng1999policy} to encourage revealing more of the visible cube pixels and the SOI. 
The reward function is as follows:
\begin{equation}
r_{t} = \lambda_{r} \Delta A^{\text{rigid}}_{t} + \lambda_{d} \Delta N^{\text{ring}}_{t} + \lambda_{f} \mathbf{I}( A^{\text{rigid}}_{t}, A^{\text{ring}}_{t}, t )
\end{equation}
where $\Delta A^{\text{rigid}}_{t} = A^{\text{rigid}}_{t} - A^{\text{rigid}}_{t-1}$ is the increased visibility of rigid object pixels in the image captured by the active camera. $\lambda_{r}$ is a scale compensation factor depending on the relative distance between the object and the camera. $\Delta N^{\text{ring}}_{t} = N^{\text{ring}}_{t} - N^{\text{ring}}_{t-1}$ is the increased number of visible deformable SOI vertices in the active camera view. $\lambda_{d}$ is a constant scale factor for the deformable object. $\mathbf{I}(\cdot)$ is an indicator function that checks if the IP policy has reached the final goal. Specifically, $\mathbf{I}(\cdot)$ returns 1 only if $A^{\text{rigid}}_{t}$ and $A^{\text{ring}}_{t}$ both reach the preset visibility thresholds before timeout. $\lambda_{f}$ is a time-dependent scale factor, defined as $100 \times (T_{\text{max}} - t)$, encouraging the IP policy to finish the task as quickly as possible.
To demonstrate the effectiveness of DAVS in comparison to the state-of-the-art, we choose a model-free RL algorithm PPO ~\cite{schulman2017proximal} as the policy search algorithm. We provide quantitative evaluations regarding the finishing episode length and the final reward, and qualitative results illustrating the learned policy behaviors among different methods.

\subsection{Baseline Methods}
We choose the PPO \cite{schulman2017proximal} and a straightforward Visual Servoing (VS) approach that focuses on maximizing the visibility of the SOI within the camera’s view as our baselines. The 2D depth and heatmap images are concatenated, and embedded as a fix-length feature vector through a convolution neural network architecture from ~\cite{mnih2015human}. The image feature vector and the end-effector location vector are concatenated and used as PPO input. The output camera and end-effector actions thus depend on both current image features and end-effector position. 

\blue{
\begin{table*}[t]
    \centering
    \vspace{9pt}
    \begin{adjustbox}{max width=0.8\linewidth}
    \setlength{\tabcolsep}{4pt}
         \begin{tabular}{llcccc}
            \toprule
            & & \multicolumn{4}{c}{\textbf{Episode Length ($\downarrow$) / Reward ($\uparrow$) / Success Rate(\%) ($\uparrow$)}}\\ 
            \cmidrule(l){3-6}
             Dynamics & Method & Fix Cam, Fix EE & Rand Cam, Fix EE & Fix Cam, Rand EE & Rand Cam, Rand EE\\ \midrule
            \multirow{2}{*}{Seen} & w/o DAVS & 200.0 / 0.3 / 0.0 & 104.4 / 9561.8 / 66.0 & 164.7 / 3537.0 / 23.0 & 110.5 / 8958.9 / 49.0\\ 
            & w/ DAVS (ours) & \textbf{39.0} / \textbf{16107.1} / \textbf{100.0} & \textbf{64.0} / \textbf{13600.7} / \textbf{84.0} & \textbf{35.24} / \textbf{16483.2} / \textbf{95.0}& \textbf{90.1} / \textbf{10998.6} / \textbf{61.0}\\ 
            \midrule
            \multirow{2}{*}{New Dynamics} & w/o DAVS & 200.0 / 0.2 / 0.0 & 116.4 / 8359.0 / 59.0 & 152.0 / 4796.1 / 30.0 & 114.5 / 8557.1 / 46.0\\ 
            & w/ DAVS (ours) & \textbf{54.4} / \textbf{14565.6} / \textbf{91.0} & \textbf{84.6} / \textbf{11541.6} / \textbf{73.0}& \textbf{39.7} / \textbf{16035.9} / \textbf{94.0} & \textbf{80.7} / \textbf{11943.3} / \textbf{66.0}\\
            \midrule
            \multirow{2}{*}{New Shapes} & w/o DAVS & 188.8 / 1125.0 / 11.9 & 122.5 / 7750.1 / 50.5 & 95.6 / 10439.2 / 70.5 & 101.5 / 9859.9 / 58.8 \\ 
            & w/ DAVS (ours) & \textbf{97.9} / \textbf{10213.0} / \textbf{66.4} & \textbf{98.2} / \textbf{10184.4} / \textbf{63.1}& \textbf{38.7} / \textbf{16141.4} / \textbf{93.4} & \textbf{72.7} / \textbf{12734.3} / \textbf{80.6}\\ 
            \bottomrule
        \end{tabular}
    \end{adjustbox}
    \caption{Performance across varied Camera and End-Effector Settings}
    \label{tab:sim_generalibility}
    \vspace{-0.5cm}
\end{table*}
}

\begin{table}[t]
    \centering
    \vspace{9pt}
    \begin{adjustbox}{max width=\linewidth}
    \setlength{\tabcolsep}{4pt}
        \begin{tabular}{
            l
            S[table-format=3.2]@{ / }S[table-format=5.2]@{ / }S[table-format=3.2]
            S[table-format=3.2]@{ / }S[table-format=5.2]@{ / }S[table-format=3.2]
            c
            }
            \toprule
            & \multicolumn{7}{c}{\textbf{Episode Length ($\downarrow$) / Reward ($\uparrow$) / Success Rate(\%) ($\uparrow$)}} \\
            \cmidrule(l){2-8}
            Method & \multicolumn{3}{c}{\textbf{CubeBagClean}} & \multicolumn{3}{c}{\textbf{CubeBagObst}} & \textbf{Total(\%) ($\uparrow$)} \\ \midrule
            w/o DAVS & 137.6 & 6248.0 & 36.0 & 87.2 & 11539.5 & 65.0 & 50.5 \\ 
            DAVS$_{\omega=0}$ & \textbf{32.0} & \textbf{16824.8} & \textbf{99.0} & 143.8 & 5739.8 & 36.0 & 67.5 \\ 
            DAVS$_{\omega=0.5}$ & 59.6 &   13950.7 & 82.0 & 90.4 & 10760.6 & 59.0 & 71.5\\ 
            DAVS$_{\omega=1}$   & 62.7 &   13740.9 & 78.0 & \textbf{68.3} & \textbf{13536.2} & \textbf{75.0} & \textbf{76.5} \\ 
            VS & 35.4 & 16107.7 & 96.0 & 137.5 & 5348.4 & 34.0 & 65.0 \\ 
            \bottomrule
        \end{tabular} 
    \end{adjustbox}
    \caption{Performance by Different Approaches}
    \label{tab:sim_compare}
    \vspace{-1cm}
\end{table}

\begin{figure*}[htbp]
  \begin{center}
    \includegraphics[width=0.9\linewidth]{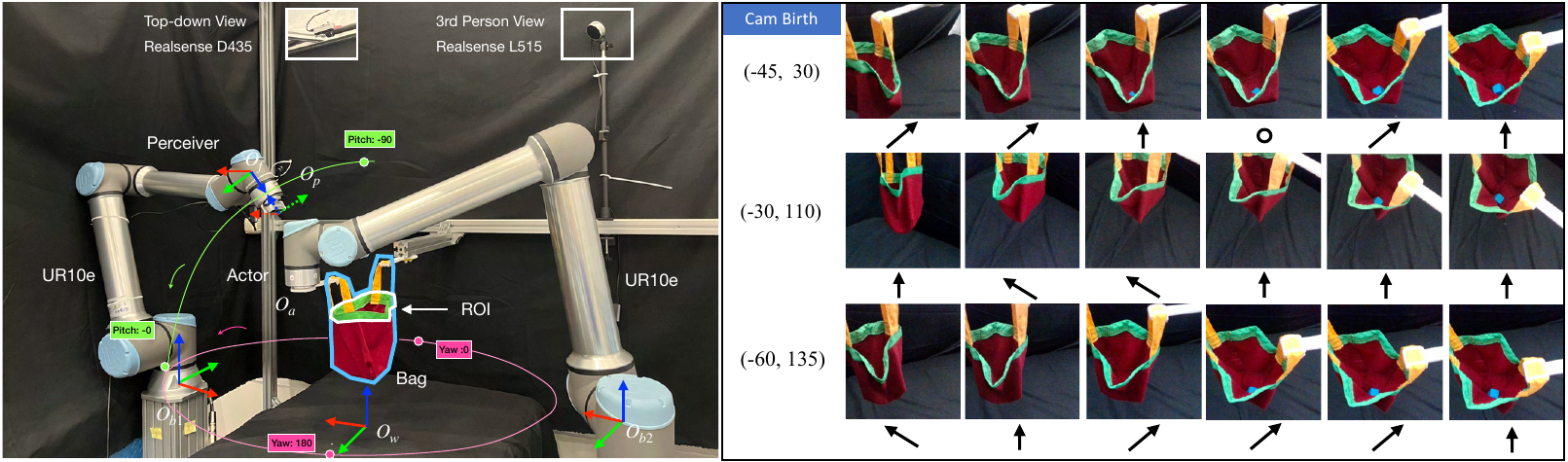}
  \end{center}
  \vspace{-0.5cm}
  \caption{
  (Left) The experimental set-up of active perception for deformable object manipulation;
  (Right) The real-world experimental results with different camera birth modes. Arrows $\uparrow$ beneath the images illustrate how the camera moves.
  }
  \label{fig_real_exp}
  \vspace{-0.5cm}
\end{figure*}

\subsection{Result}
\subsubsection{CubeBagClean}
Fig. \ref{exp2_clean_obst} shows the learning process on 4 subtasks, comparing PPO without DAVS to the variant with the proposed DAVS on the metrics of the finished episode length and total discounted rewards. It is evident that the performance of the baseline is improved by a significant margin by introducing DAVS (with manifold), achieving lower episode length and higher reward in all four subtasks. This indicates that DAVS successfully encodes the coordination between the camera and end-effector agents and expedites the finding of better solutions. We also visualize an example of the learned IP policies in Fig. \ref{case_qualitative}. The last row of the figure illustrates the dynamic evolution of the manifold in a CubeBagClean episode in the first row. The camera position, SOI, projected SOI, and DAVS are plotted in different colors.  For each subtask, we report the episode length, total reward, and the success rate on 100 random scenes as presented in Table \ref{tab:sim_generalibility}.

\subsubsection{CubeBagObst}
The additional circular obstacle in this scenario introduces more view occlusion and hence requires more complex maneuver and coordination of camera and end-effectors.
In this experiment, besides a default PPO with and without DAVS, we also take a PPO with static vision into account for comparison to show the insufficiency of static vision methods. From Fig. \ref{exp2_clean_obst}, we observe the agents in IP with DAVS outperform the counterparts without DAVS or with the static vision solution, particularly when the camera starts from a fixed but challenging side perspective. The qualitative result is shown in Fig. \ref{case_qualitative}.

\subsubsection{Generalization to objects with new properties}

To evaluate the trained networks' generalizability, we analyzed their performance when interacting with bags exhibiting unseen yet similar dynamics, despite the model being trained with a single-bag dynamic setup. Specifically, we explored the impact of random variations in the bag's spring elastic stiffness and damping stiffness on its dynamics. 

For each CubeBagClean subtask, we created 100 scenes with random bag dynamics to evaluate the models' performance, with and without DAVS. As presented in Table \ref{tab:sim_generalibility}, the IP policy, when utilizing DAVS, not only consistently outperforms its counterpart without DAVS but also demonstrates superior generalizability, underscoring the robustness and effectiveness of our methodology.

Furthermore, we investigated the impact of variations in shape, which were not seen during training. Specifically, we created scenes with 36 bags with shapes adapting from \cite{antonova2021dynamic}. As shown in Table \ref{tab:sim_generalibility}, the IP policy with DAVS performs better. We observed that certain tasks with unseen shapes were easier to achieve due to two reasons: 1) some new bags had large openings, making it easier to capture the object inside during manipulation; 2) some new bags were more shallow, causing the contained cube to be closer to the camera. Consequently, the rigid cube area appeared relatively larger than in the original setup in the captured image, making reaching the pre-set cube area threshold easier.

\subsubsection{Comparison with other baselines}
As shown in Table \ref{tab:sim_compare}, with the reduction of $\omega$ to shrink the exploration space for camera actions, the camera is more likely to move directly into the manifold centroid. The PPO policy with DAVS achieves shorter episode lengths, higher total rewards, and higher success rates.
While $\boldsymbol{\omega}=0$ significantly improves performance in simpler scenarios (CubeBagClean), higher values of $\omega$ (DAVS$_{\omega=0.5}$ and DAVS$_{\omega=1}$ ) are crucial for handling more complex environments (CubeBagObst).
The Visual Servoing method (VS) performs better than DAVS$_{\omega=1}$ in some aspects but is outperformed by DAVS with $\omega$ settings on CubeBagObst. 
In general, DAVS$_{\omega=1}$ achieves the best overall balance, making it the most robust approach among the tested methods. We also examined the failure cases in the simulation experiments. In our Pybullet simulation setup, we executed the movements of the camera and the actor simultaneously. Most failures occurred when the simulated gripper moved too quickly, causing overstretching of the deformable handles. Additionally, the relatively slow movement of the camera made it more difficult to locate the object, especially in scenes involving obstacles. Furthermore, under the challenging generalization test, the IP method failed particularly when the stiffness differed significantly or with a small bag opening.

\section{Real-robot Experiments}
To assess the effectiveness of the policy trained using DAVS for 
deformable object manipulation with active perception, we selected a policy fine-tuned within the CubeBagClean simulation environment (featuring random camera and end-effector positions) for real-world experimentation.
Figure \ref{fig_real_exp} illustrates our real-world experimental configuration for active perception in deformable object manipulation tasks. Here, a UR10e robotic arm (the actor) manipulates one handle of a fabric bag (with the other handle fixed) using a 3D-printed end-effector. Concurrently, a second UR10e robotic arm (the perceiver) equipped with a RealSense D435 depth camera observes the environment.
We manually measured the transformation matrices between the world coordinate frame $O_w$ and the perceiver's base frame $O_{b1}$, as well as between $O_w$ and the actor's base frame $O_{b2}$.
Mirroring the camera perceiver configuration from our PyBullet training environment, the camera's action space is limited to a hemispherical area, with its orientation constantly aimed at the sphere's center—which also coincides with the origin of the world frame $O_w$. Given the known central point of the camera's field of view and the sphere's radius, the camera's position and orientation are straightforwardly defined by its pitch and yaw angles. Taking into account the kinematic constraints, these angles are set within the ranges of [-89, -30] degrees for pitch and [45, 135] degrees for yaw.
The inputs to our real RL agent comprise the end-effector pose, depth image and a ring-shaped heat map. At each step, the DAVS generates a construction based on the captured green Region of Interest (ROI) of the fabric bag, which is defined by the bag's green opening rim. This green rim is roughly identified by an overhead camera and then transformed into the world frame using a fixed transformation. The outputs generated are the pitch, yaw, and relative motion actions of the actor.
Following approximately 5,000 episodes of fine-tuning, the perceiver and actor synergistically cooperate to reliably locate a blue cube within the bag. Figure \ref{fig_real_exp} showcases the successful outcomes of these real-world experiments.

\section{Conclusion}
\label{sec:conclusion}
In this work, we address deformable object manipulation through Interactive Perception in a system involving an active camera, a robotic end-effector, and a deformable object. We contribute to a novel formulation of the problem with common model assumptions and enable tractable computation through our proposed framework. Then, we perform simulation experiments with different view occlusion situations and demonstrate that our proposed framework with DAVS outperforms the state-of-the-art methods. Finally, we confirm that our method generalizes well to bags featuring previously unseen dynamical properties, as well as their efficacy within an actual real-world interactive perception system. 
The challenges presented in this work lead to the future directions of improving hand-eye coordination, such as explicitly incorporating dynamics prediction within the IP framework, and more effective parameterized DAVS constructions with task dependency for deformable object manipulation.

\vspace{-0.2cm}
\bibliographystyle{IEEEtran}
\bibliography{ref}

\end{document}